\documentclass[letterpaper, 10 pt, journal, twoside]{IEEEtran}

\usepackage{url}
\usepackage{amssymb}
\usepackage{graphicx}
\usepackage{algorithm,algorithmic}	
\usepackage{tikz}
\usepackage{mathtools}
\usepackage[compress]{cite}
\usepackage{amsmath}
\usepackage{dsfont}
\usepackage{xfrac}

\newtheorem{assumption}{Assumption}
\newtheorem{theorem}{Theorem}

\newcommand{\argmax}{\mathop{\text{arg\,max}}}
\newcommand{\trace}{\text{trace}}
\newcommand{\supp}{\text{supp}}

\DeclareMathOperator{\st}{s.\hspace{-.03in}t.}

\title{Optimal Stochastic Evasive Maneuvers Using the Schr\"{o}dinger's Equation }

\author{
Farhad Farokhi\thanks{Corresponding author. e-mail: farhad.farokhi@data61.csiro.au} and Magnus Egerstedt\thanks{F. Farokhi is with the CSIRO's Data61 and  the University of Melbourne. His work was supported by the McKenzie Fellowship from the University of Melbourne and the VESKI Victoria Fellowship from the Victorian State Government. M. Egerstedt is with the Institute for Robotics and Intelligent Machines, Georgia Institute of Technology. His work was supported by National Science Foundation. }
}

\begin{document} 

\maketitle 
\thispagestyle{empty}

\begin{abstract}
In this paper, preys with stochastic evasion policies are considered. The stochasticity adds unpredictable changes to the prey's path for  avoiding predator's attacks. The prey's cost function is composed of two terms balancing the unpredictability factor (by using stochasticity to make the task of forecasting its future positions by the predator difficult) and energy consumption (the least amount of energy required for performing a  maneuver). The optimal probability density functions of the actions of the prey for trading-off unpredictability and energy consumption is shown to be characterized by the stationary Schr\"{o}dinger's equation.
\end{abstract}


\section{Introduction}
\IEEEPARstart{O}{ne} aspect of evasion for preys in nature is to deceive predators in terms of their escape trajectory. The prey can utilize randomness in its decision making process to ensure that the predator cannot anticipate its future positions on the escape trajectory accurately, contrary to the case that the prey always escapes directly away from the predator~\cite{CARD2012180}. In nature, this behaviour is demonstrated by mantises, moths, and green lacewings performing sudden, unpredictable changes in flight path to avoid attacks of bats~\cite{yager1990ultrasound}. Cockroaches also randomly select their escape bearing from a set of possible trajectories at fixed angles away from the threat~\cite{domenici2008cockroaches}. The randomness in the escape trajectory seems to be shared by many other insects~\cite{domenici2011animal}. The aim of this paper is to draw inspiration from these preys to devise stochastic evasive maneuvers for engineering applications. 

Investigation of evasive maneuvers is an active area of research in robotics and aerospace engineering; see~\cite{ho1965differential,shima2002time, oyler2016pursuit,casini2017improved,shinar1980three} spanning over five decades. In these studies,  the underlying methodology for evasion is typically deterministic. Analysing the dynamics of the predator, specifically its constraints and short comings, such as maximum speed and turning radius, reveals a deterministic path for evasion (based on mechanical and geometric advantages). This  sometimes results in a game-theoretic approach for decision making, dubbed as pursuit-evasion games~\cite{isaacs1999differential}. The application and analysis of these games do not however end with engineering~\cite{miller1994protean,doi:10.1093/icb/ict061,doi:10.1093/icb/icv027,dugatkin2000game}.

Fundamental differences between theoretical investigation of optimal evasion trajectories and the behaviour of preys in nature was also observed within the biological studies~\cite{domenici2011animal_1}, where it was noted that the theoretical studies of optimal evasion trajectories often fall back on the relative speeds of predator and prey~\cite{Arnott193,domenici2002visually,WEIHS1984189} and do not accommodate one of the main properties postulated for evasion trajectories, namely their unpredictability, which is fundamental for preventing predators from learning a repeated pattern of prey response~\cite{comer2009behavioral,godin1997behavioural,humphries1970protean}. 

\section{Modelling}
Consider a prey that has been detected or observed at an initial position $y_0\in\mathbb{R}^p$, where $p=2$ for planar and $p=3$ for aerial preys, at time $t_0=0$ by a predator. The dynamics of the prey is given by 
\begin{subequations}
\label{eqn:main:system}
\begin{align} 
\dot{x}(t)&=f_t(x(t),u(t)), \quad x(t_0)=x_0,\\
y(t)&=g_t(x(t)),
\end{align}
\end{subequations}
where $x(t)\in\mathbb{R}^n$ is the state of the prey (e.g., its position, velocity, and orientation), $u(t)\in\mathbb{R}^m$ is the control input, and $y(t)\in\mathbb{R}^p$ is the output measurements, which is the position in this case. Note that, due to consistency, $y_0=g_{t_0}(x_0)$. In what follows, the dynamical system~\eqref{eqn:main:system} is assumed to be controllable in the sense that there exists a sequence of actions for it to be able to get from any state to another state within a non-zero horizon. Controllability could be a reasonable assumption as otherwise the prey cannot perform necessary maneuvers to escape, potentially rendering it vulnerable. Although any state is reachable, some maneuvers consume more energy and may not be practically feasible.

The predator is assumed to be interested in estimating the position of the prey in $T$ seconds, at time instant $t_1=t_0+T$. This could be motivated by that the predator aims to intercept the prey at that position. By anticipating the trajectory of the prey, the predator can gain an advantage and capture the prey easier. For instance, snakes anticipate future behaviour of their prey and strike where they would be later~\cite{catania2009tentacled}. Other predators have been also noted to actively predict future movements of their preys~\cite{borghuis2015role,naturedragonfly}.

At $t_1$, the state of the prey is $x_1=x(t_1)$. The position of the prey $y_1=g_{t_1}(x_1)$ is determined by the control actions of the prey over the window of time of interest $[t_0,t_1]$ denoted by $(u(t))_{t=t_0}^{t_1}$. The problem of interest for the prey is to determine the position $y_1$ to  balance between energy consumption and its ability to evade (in the sense of making the predator's estimation error relatively large to remain unpredictable). 

In order to accomodate unpredictability of the prey along its evasion trajectories, which is postulated to be a fundumental aspect of evasion in nature~\cite{comer2009behavioral,godin1997behavioural,humphries1970protean, domenici2011animal}, the prey selects the desired position $y_1$ according to a randomized policy captured by a probability density $\xi_{x_0}$. This implies that
$\mathbb{P}\{y_1\in\mathcal{Y}\}=\int_{y'_1\in\mathcal{Y}} \xi_{x_0}(y'_1)\mathrm{d}y'_1$ for any Lebesgue-measurable set $\mathcal{Y}\subseteq\mathbb{R}^p$. Note that $\xi_{x_0}$ is not a conditional probability function as $x_0$ is not necessarily a random variable (there is no assumption regarding a prior for $x_0$). The notation $\xi_{x_0}(\cdot)$ emphasizes possible dependency of the probability density function of $y_1$ on $x_0$ as a parameter\footnote{As observed in the remainder of the paper, in some cases, the optimal policy does not depend on $x_0$. }. For instance, at high altitudes, downward dives might be more likely. The task at hand, in this paper, is to find $\xi_{x_0}$ to  balance between evasion efficiency and energy consumption. 

Since the prey is controllable, there always exist control actions $(u(t))_{t=t_0}^{t_1}$ to bring the prey from the position $y_0$, with a given initial state of $x_0$, to the position $y_1$ over the horizon $[t_0,t_1]$. The minimum required energy for the maneuver is 
\begin{subequations}
\begin{align}
U_{x_0}(y_1):=\min_{
\begin{array}{c}
\scriptstyle u(t)\in\mathbb{R}^m,\forall t\in[t_0,t_1]\\
\scriptstyle x(t)\in\mathbb{R}^n,\forall t\in[t_0,t_1]
\end{array}
} &\;\int_{t_0}^{t_1} \ell_t(x(t),u(t))\mathrm{d}t,\\
\st\hspace{.42in}& \dot{x}(t)=f_t(x(t),u(t)),\nonumber\\
&\hspace{0.4in} \forall t\in[t_0,t_1], \\
& x(t_0)=x_0,\\
& y_1=g_{t_1}(x(t_1)),
\end{align}
\end{subequations}
where $\ell_t:\mathbb{R}^n\times\mathbb{R}^m\rightarrow\mathbb{R}_{\geq 0}$ relates the energy consumption of the prey to its control input and state. Here, $\mathbb{R}_{\geq 0}:=\{x\in\mathbb{R}\,|\,x\geq 0\}.$ To keep its energy consumption low, the prey can select $y_1$ so that $U_{x_0}(y_1)$ is kept  low.  Since the prey's policy is stochastic, it keeps the following average cost low:
\begin{align}
\psi(\xi_{x_0}):=&\int_{y_1\in\supp(\xi_{x_0})}U_{x_0}(y_1)\xi_{x_0}(y_1)\mathrm{d}y_1\nonumber\\=&\mathbb{E}\{U_{x_0}(y_1)\},
\end{align}
where $\supp(\xi_{x_0}):=\{y_1\in\mathbb{R}^p\,|\,\xi_{x_0}(y_1)>0\}$ is  the support set of the density function $\xi_{x_0}$. 

Another facet of evasion is to avoid heading back towards the predator, even when reacting unpredictably. In fact, it is desired to head away from the predator so that the evasion is complete. Thus the preferred directions head away from the predator but with enough choices for unpredictable maneuvers. To capture this, define the set 
\begin{align}
\aleph_{x_0}:=
\{y_1\in\mathbb{R}^p\,|\,&\mbox{angle between $y_1-y_0$}\nonumber \\
&\mbox{and $y_0-z_0$ is smaller than $\theta_{\max}$} \},
\end{align}
where $z_0$ is the position of the predator at time $t_0$ and $\theta_{\max}\in(0,\pi]$.  The search is restricted to the set of probability density functions that $\supp(\xi_{x_0})\subseteq \aleph_{x_0}$, i.e., the set of probability density functions $\xi_{x_0}$ that, with probability one, pick $y_1$ to head away from the predator  (according to the property that the angle between $y_1-y_0$ and $y_0-z_0$ is smaller than or equal to $\theta_{\max}$). In what follows, $\partial \aleph_{x_0}$ is the boundary of~$\aleph_{x_0}$.

Effective tracking of the prey improves the chances of the predator to capture it. The predator's prediction of the location of the prey is denoted by $\hat{y}_1$.  A measure of the effectiveness of the predator's ability to predict the behaviour of the prey is $\mathbb{E}\{\|\hat{y}_1-y_1\|_2^2\}$. The objective of the prey is to ensure that $\mathbb{E}\{\|\hat{y}_1-y_1\|_2^2\}$ is maximized (albeit in balance with other factors such as energy consumption to ensure the policy is implementable), irrespective of the predator's policy for determining $\hat{y}_1$. The independence from the predator's policy is motivated by that the prey does not know the estimation policy of new predators or that it might want to be prepared for the worst-case outcome. 

The following assumptions make the problem of searching for this policy more tractable. 
\begin{assumption} \label{assum:2} (\textit{i}) $\xi$ is twice continuously differentiable, (\textit{ii}) $\{y_1\in\aleph_{x_0}\,|\,\xi_{x_0}(y_1)=0\}$ has zero Lebesgue-measure for all $x_0$ and $\xi_{x_0}(y_1)=0$ for all $y_1\in\partial \aleph_{x_0}$, and (\textit{iii}) $\mathbb{E}\{y_1\}$ is bounded for all $x_0$.
\end{assumption}

Assumption~\ref{assum:2}~(\textit{i}) simplifies the search for the optimal policy $\xi_{x_0}$ by allowing the use of calculus of variation~\cite{kirk2012optimal}.
Assumption~\ref{assum:2}~(\textit{ii}) ensures that the Fisher information matrix, in Theorem~\ref{tho:1}, is well-defined, and is inversely related to the estimation error of the predator, and that its trace is a convex function of the density function $\xi_{x_0}$. Finally, in the absence of Assumption~\ref{assum:2}~(\textit{iii}), the average energy required for realising $y_1$ is unbounded and thus such a policy cannot be physically realized. This paper adopts the notation $\nabla_{x}f(x)$ to denote the gradient of a continuously differentiable mapping $f(x)$, which is assumed to be a column vector.

\begin{theorem} \label{tho:1} Under Assumption~\ref{assum:2}, it can be shown that
\begin{align} \label{eqn:lowerbound}
\mathbb{E}\{\|\hat{y}_1-y_1\|_2^2\}\geq 1/\trace(\mathcal{I}(\xi_{x_0})),
\end{align}
where $\mathcal{I}(\xi_{x_0})$ is the Fisher information associated with the density function $\xi_{x_0}$ defined as
\begin{align} \label{eqn:def:Fisherinformation}
\mathcal{I}(\xi_{x_0}):=&\int_{y_1\in\supp(\xi_{x_0})} \nabla_{y_1} \log(\xi_{x_0}(y_1)) \nonumber\\&\hspace{.3in}\times\nabla_{y_1} \log(\xi_{x_0}(y_1))^\top \xi_{x_0}(y_1)\mathrm{d}y_1.
\end{align} 
\end{theorem}

\begin{IEEEproof} See Appendix~\ref{proof:tho:1}.
\end{IEEEproof}

Note that, strictly speaking,~\eqref{eqn:def:Fisherinformation} is not the classic definition of the Fisher information matrix; see, e.g.,~\cite{cramerraotheorem_science}. This can be observed by noting that in the definition of the Fisher information matrix the variables with respect to which derivatives and integrations are performed are not the same, contrary to~\eqref{eqn:def:Fisherinformation}. This is due to the fact that, in the Cram\'{e}r-Rao bound, a randomized measurement is first realized and from that some deterministic parameters are estimated. However, in this paper, a deterministic measurement (regarding the initial position of the prey) is first taken and then a randomized movement in the state is realized with the ultimate aim being to estimate the-said random state and not the deterministic initial condition. Forgetting about this ``philosophical'' difference, the proof techniques are similar to that of the Cram\'{e}r-Rao bound with subtle,  yet important, differences. Due to these  similarities, the name Fisher information matrix is adopted for $\mathcal{I}(\xi_{x_0})$. The bound in~\eqref{eqn:lowerbound} is tight in the sense that there exist $\xi_{x_0}$ and $\hat{y}_1$ such that~\eqref{eqn:lowerbound} holds with equality\footnote{Let $\xi_{x_0}$ be a Gaussian distribution with zero mean and unit variance (for which $\trace(\mathcal{I}(\xi_{x_0}))=p$) and set $\hat{y}_1$ to be the maximum likelihood estimator (i.e., $\hat{y}_1=\argmax_{y_1}\xi_{x_0}(y_1-y_0)=y_0$ implying that $\mathbb{E}\{\|\hat{y}_1-y_1\|_2^2\}=p$). Then, $\mathbb{E}\{\|\hat{y}_1-y_1\|_2^2\}=1/\trace(\mathcal{I}(\xi_{x_0}))$ if $p=1$.}. 

The problem of finding the optimal policy $\xi_{x_0}$ for striking a balance between evasion efficiency and energy consumption while running away from the predator can be posed as an infinite-dimensional optimization problem in
\begin{subequations}\label{main:eqn:problem_formulation}
\begin{align} 
\min_{\xi_{x_0}\in \Xi} & \;\trace(\mathcal{I}(\xi_{x_0}))+\varrho\psi(\xi_{x_0}),\\
\st \;& \;\supp(\xi_{x_0})\subseteq \aleph_{x_0}.
\end{align}
\end{subequations}
where $\Xi$ is the set of all policies that admit Assumption~\ref{assum:2} and $\varrho>0$ is a constant balancing the trade-off between evasion efficiency and energy consumption. As $\varrho$ increases, the emphasis on energy consumption also increases (i.e., the prey acts more sedentarily). 

\begin{figure}
\centering
\includegraphics[width=1.0\linewidth]{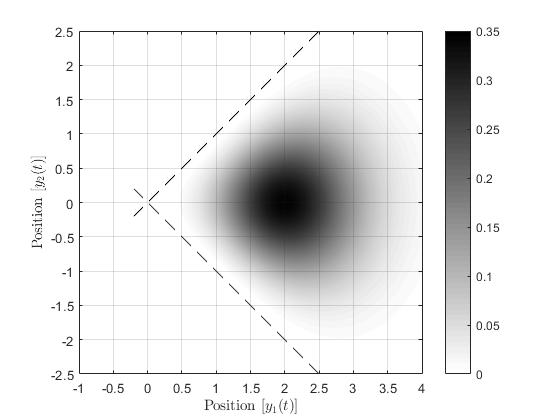}
\caption{\label{fig:1} The optimal privacy preserving $\xi_{x_0}$ for a single-integrator robot moving on the ground. The dashed lines show the boundary of the set $\aleph_{x_0}$. The probability density function $\xi_{x_0}$ takes larger values at darker areas and thus the prey is more likely to appear at those points at time $t=t_1$. }
\end{figure}

\section{Main Results} 
The following theorem captures the optimal stochastic evasive maneuver, in the sense of the solution of~\eqref{main:eqn:problem_formulation}. It shows that the optimal probability density functions of the
actions of the prey is characterized by the
Schr\"{o}dinger's equation.

\begin{theorem} \label{main:results}
The solution of~\eqref{main:eqn:problem_formulation} is given by 
\begin{align*}
\xi_{x_0}(y_1)
=\begin{cases}
u_{x_0}(y_1)^2, & y_1\in\aleph_{x_0},\\
0, & y_1\notin \aleph_{x_0},
\end{cases}
\end{align*}
where $u_{x_0}(y_1)$ is determined by
\begin{subequations} 
\label{main:eqn:tho:nonlinear_query}
\begin{align} 
&\nabla^2_{y_1}u_{x_0}(y_1)+\frac{1}{4}(\mu-\varrho U_{x_0}(y_1))u_{x_0}(y_1)=0,\nonumber\\
&& \hspace{-.7in} \forall y_1\in\aleph_{x_0},\\
&\int u_{x_0}(y_1)^2\mathrm{d}y_1=1,\\
&u_{x_0}(y_1)\neq 0 && \hspace{-.7in} \forall y_1\in\mathrm{int}(\aleph_{x_0}),\\
&u_{x_0}(y_1)=0 && \hspace{-.7in} \forall y_1\in\partial(\aleph_{x_0}),
\end{align}
\end{subequations}
for some mapping $\mu$. Furthermore, all solutions (if multiple) satisfying~\eqref{main:eqn:tho:nonlinear_query} exhibit the same cost.
\end{theorem}

\begin{IEEEproof}
See Appendix~\ref{proof:main:results}.
\end{IEEEproof}

Theorem~\ref{main:results} shows that (the square root of) the prey's probability density function for selecting its destination must satisfy the time-invariant (stationary) Schr\"{o}dinger's equation. This is an interesting observation illustrating that quantum particles play an elaborate game of pursuit-evasion with measurement devices. Within the context of quantum mechanics, the same formula can be obtained by choosing Fisher information as a measure of disorder~\cite{PhysRevA414265}; however, the choice of Fisher information as a measure of disorder in quantum mechanics is philosophical while, in this paper, the choice stems from stochastic evasion. 

As an example consider a single-integrator dynamics robot moving on the ground with simple dynamics:
\begin{subequations}\label{eqn:diffwheel}
\begin{align}
\dot{x}(t)&=u(t), \quad x(t_0)=y_0,\\
y(t)&=x(t),
\end{align}
\end{subequations}
where $x(t)\in\mathbb{R}^2$ is the state (only the position for  single-integrator  robots) and $u(t)\in\mathbb{R}^2$ is the control input (the velocity in each direction). Furthermore, assume that
$\ell_t(u(t),x(t))=u(t)^\top u(t).$ For these systems, $U_{x_0}(y_1)$ can be explicitly calculated as $U_{x_0}(y_1)=(1/T)(y_1-y_0)^\top(y_1-y_0).$ Therefore, the energy required for moving from $y_0$ to $y_1$ is captured by the distance between those points. Set $y_0=0$ without loss of generality. Assume that the predator is located in $z_0=[-2 \; 0]^\top$.  We can compute the optimal policy using Theorem 2; see Appendix~\ref{appendixC} for the complete algebraic computations. Figure~\ref{fig:1} illustrates $\xi_{x_0}$ for $\theta_{\max}=45^{\circ}=\pi/4\,\mathrm{rad}$, $T=1$, and $\varrho=1$. The dashed lines show $\partial\aleph_{x_0}$. The probability density function $\xi_{x_0}$ is larger at the darker areas and thus the prey is more likely to appear at those points at time $t=t_1$. By changing the constant $\varrho$ the size of the dark region changes.

\begin{figure}
\centering
\includegraphics[width=1\linewidth]{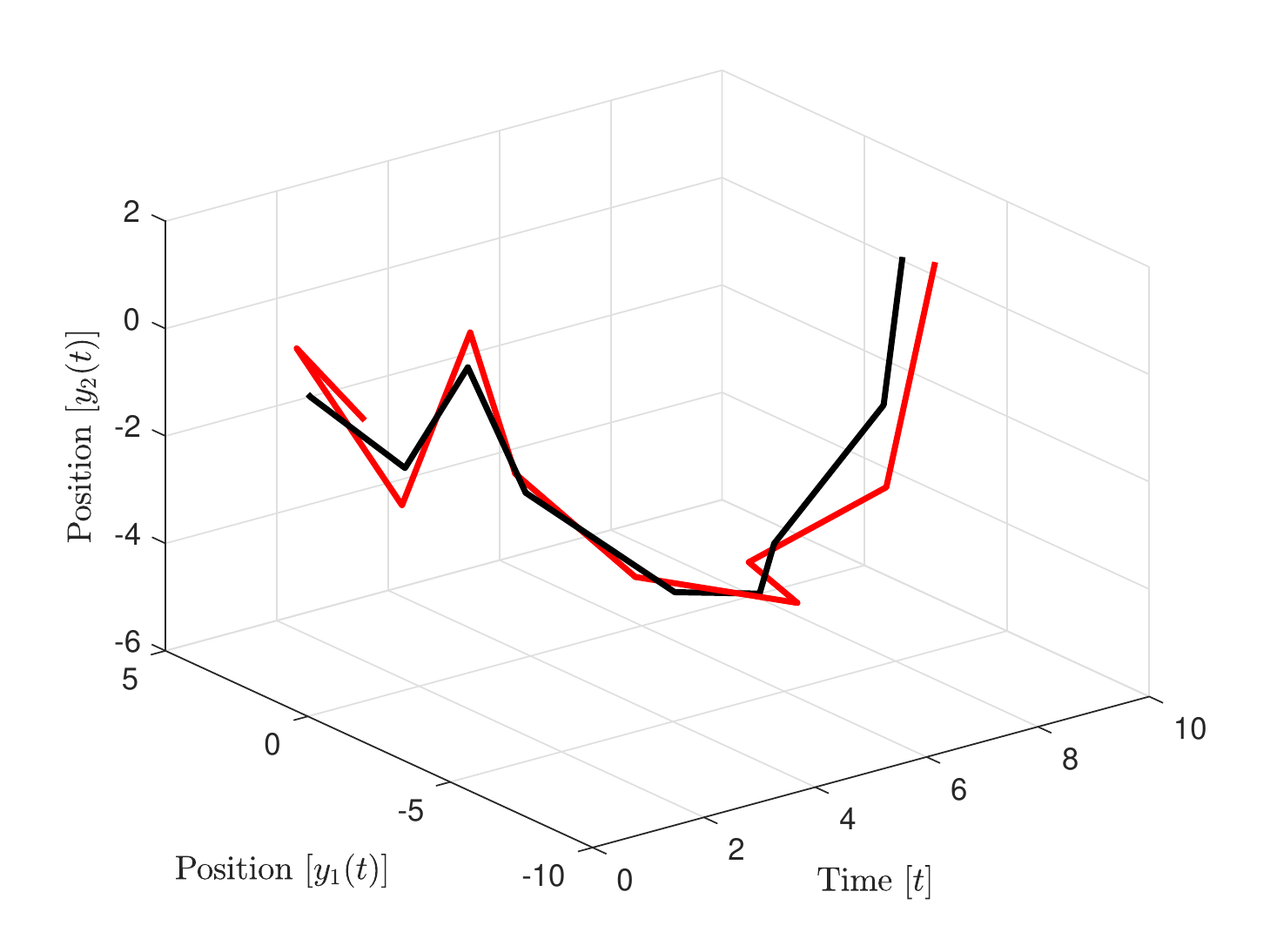}
\caption{\label{fig:4a} The trajectory of the prey when using the optimal evasion policy (black) and the trajectory of the predator (red). }
\end{figure}

\begin{figure}
\centering
\includegraphics[width=1\linewidth]{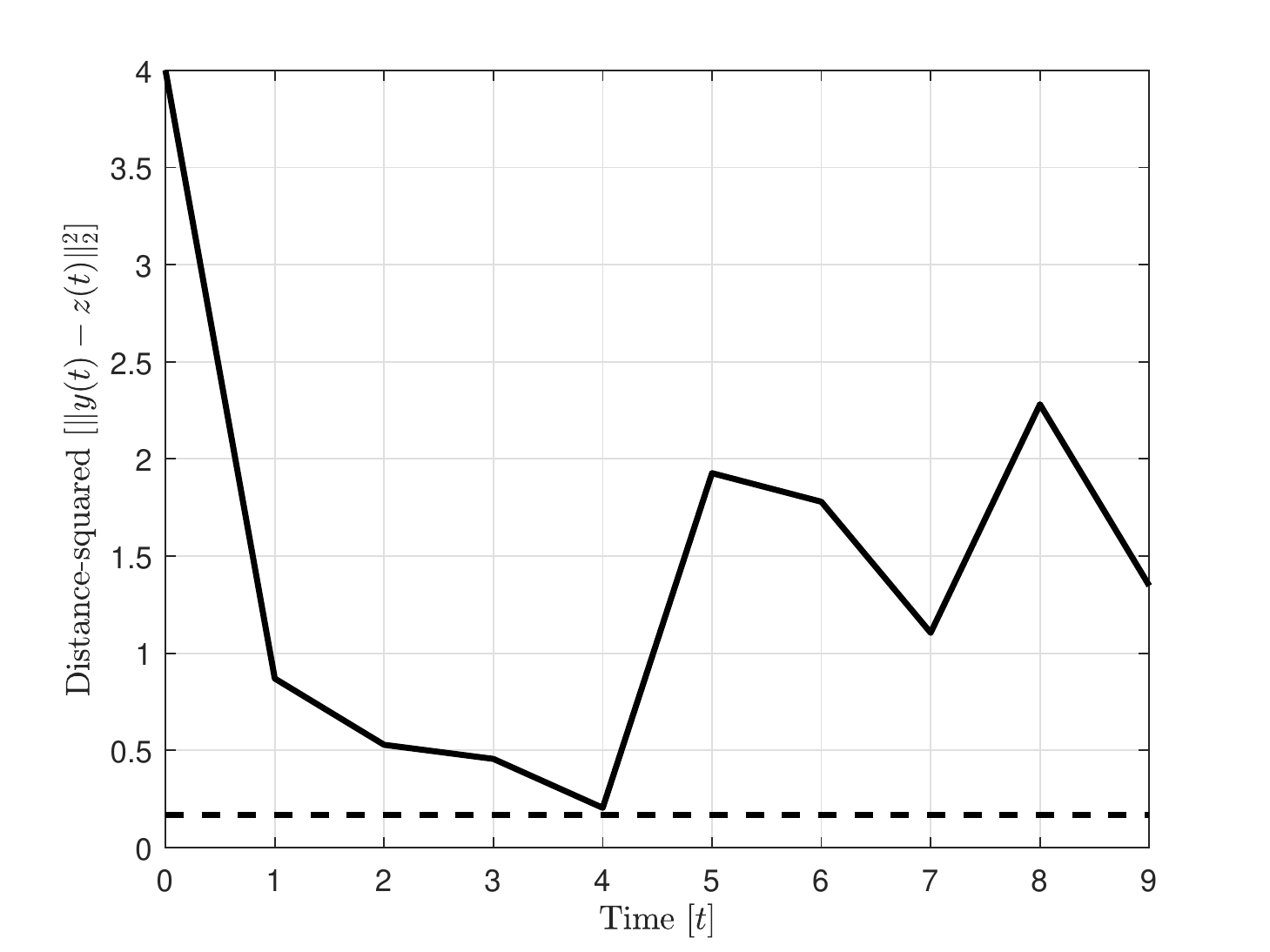}
\caption{\label{fig:4b} The distance between predator and pray (black). The predator and prey never cross path. The dashed (black) line illustrates the statistical lower bound on the distance between predator and prey given by~\eqref{eqn:lowerbound}. }
\end{figure}

\begin{figure}
\centering
\includegraphics[width=1\linewidth]{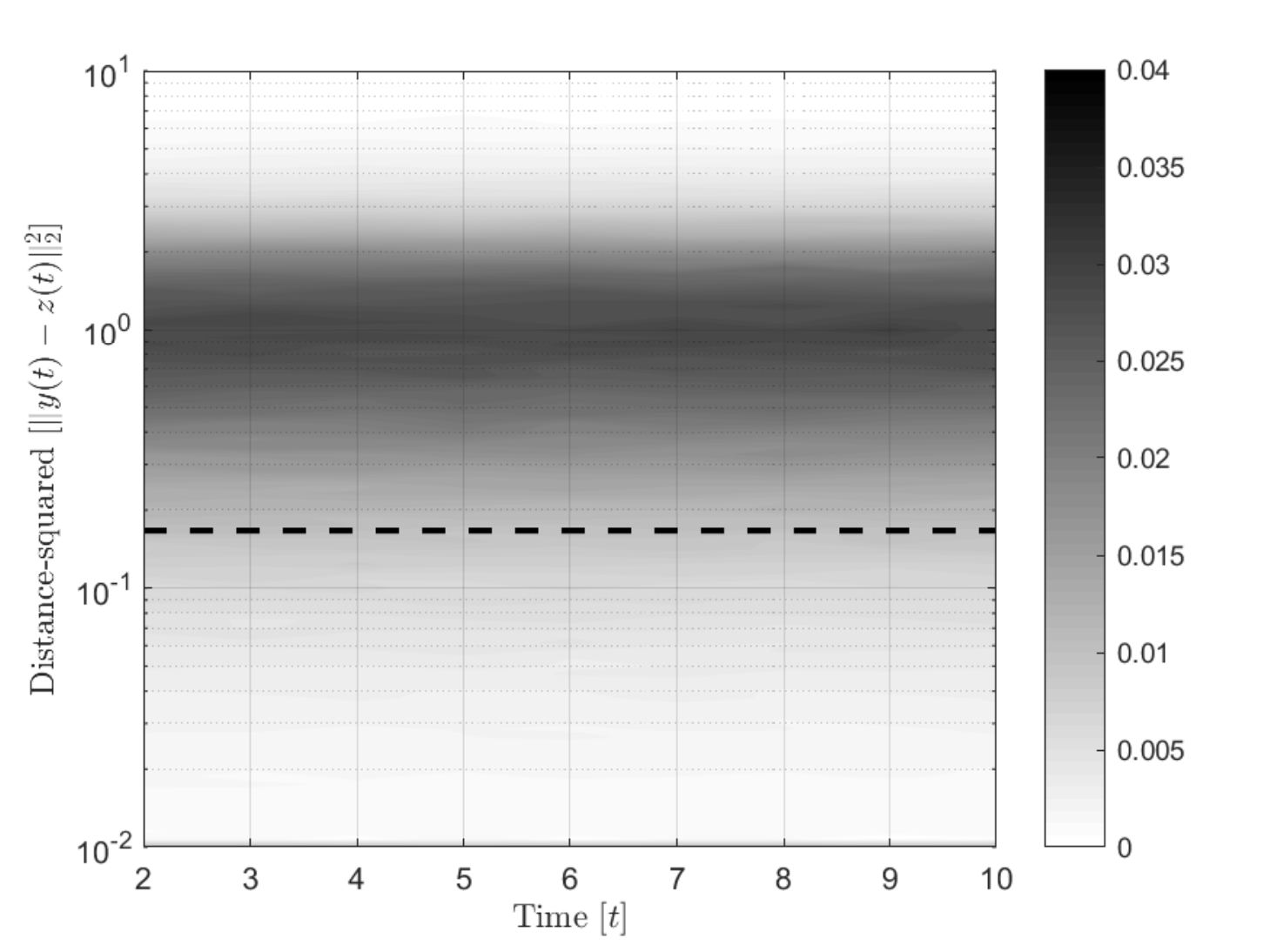}
\caption{\label{fig:5} The probability density function of the distance between predator and prey across time.  }
\end{figure}

\begin{figure}
\centering
\includegraphics[width=1\linewidth]{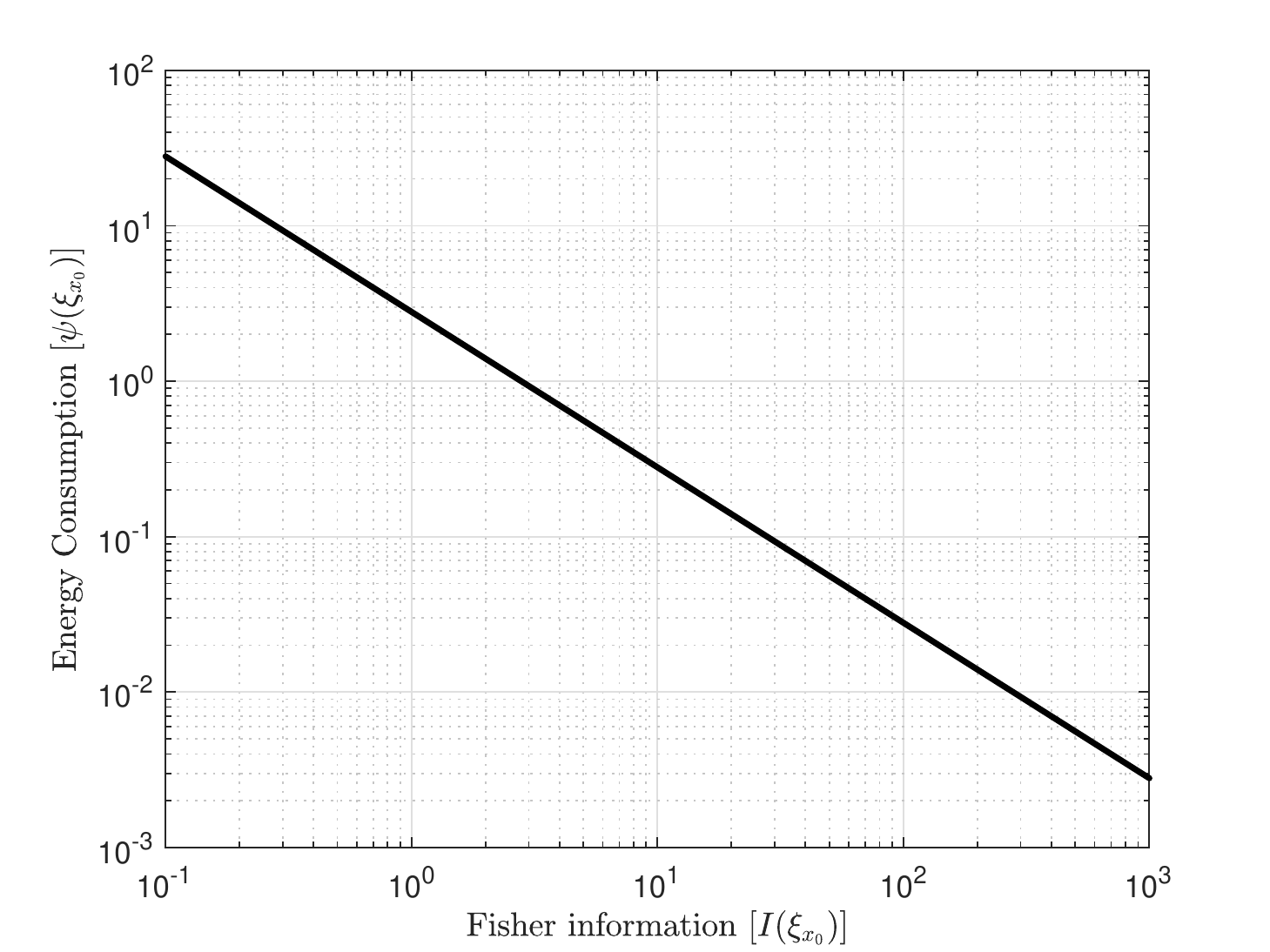}
\caption{\label{fig:6} The relationship between Fisher information $\mathcal{I}(\xi_{x_0})$ and expected energy consumption $\psi(\xi_{x_0})$. }
\end{figure}

If evader repeats the optimal policy for evasion at multiple time instances, it must follow the way points $y((k+1)T)$, which are random variables with the probability density  function  $\xi_{y(kT)}$. Note that $\xi_{y(kT)}$ is a function of $z(kT)$, which is the position of predator at time $kT$, as the set $\aleph_{y(kT)}$ changes with $z(kT)$. For this example, consider the case where the predator is also a single-integrator robot with the dynamics in~\eqref{eqn:diffwheel}. Assume that the predator uses the least mean square estimator~\cite[p.\,30]{anderson2012optimal} for estimating the prey's position as $\hat{y}((k+1)T)=\mathbb{E}\{y((k+1)T)|(y(t))_{t=0}^{kT}\}$. The predator then tries to intercept the prey by following the way points given by $z((k+1)T)=\hat{y}((k+1)T).$
Figure~\ref{fig:4a} illustrates a realization of the trajectory of the prey when using the optimal evasion policy in Theorem~\ref{main:results} (black) and the trajectory of the predator (red). The predator and the prey never cross path. To observe this, Figure~\ref{fig:4b} shows the distance between the predator and the pray (solid black) for the trajectories in Figure~\ref{fig:4a}. The dashed black line illustrates the statistical lower bound on the distance between predator and prey given by~\eqref{eqn:lowerbound}, which follows from that
$
\mathbb{E}\{\|z(kT)-y(kT)\|_2^2\}
=\mathbb{E}\{\|\hat{y}(kT)-y(kT)\|_2^2\}
\geq 1/\trace(\mathcal{I}(\xi_{x_0}))
=1/6.
$
Note that variations in the distance between the prey and predator are expected as the guarantee in~\eqref{eqn:lowerbound} only holds in expectation. Figure~\ref{fig:5} illustrates The probability density function of the distance between predator and prey across time. This density is estimated using $10^6$ randomly generated trajectories. Based on this figure, it can be seen that $\mathbb{P}\{\|z(kT)-y(kT)\|_2^2\geq 1/\trace(\mathcal{I}(\xi_{x_0}))\}\gtrapprox 0.85$. Finally, it remains to discuss the relationship between energy consumption and evasion capability. As expected, with increasing the energy consumption, the prey can implement more intensive maneuvers  to escape. Figure~\ref{fig:6} shows the relationship between Fisher information $\mathcal{I}(\xi_{x_0})$ and expected energy consumption $\psi(\xi_{x_0})$, various points on this line can be achieved with a different $\varrho$.

\section{Conclusions and Future Work}
We proved that the optimal probability density functions of the actions of the prey for trading-off unpredictability and energy consumption for implementing stochastic evasion manoeuvres is shown to be characterized by the stationary Schr\"{o}dinger's equation. Future work can focus on computing the optimal probability density function for practical robot dynamics.

\bibliographystyle{ieeetr}
\bibliography{scibib}

\appendices

\section{Proof of Theorem~\ref{tho:1}}
\label{proof:tho:1}
For all constants $(c_i)_{i=1}^p$, with $c_i\in\mathbb{R}^p$ for all $i$, it could be shown that
\begin{align}
\mathbb{E}\bigg\{&\bigg(\hat{y}_1-y_1-\sum_{i=1}^p c_i \frac{\partial}{\partial y_{1,i}}\log(\xi_{x_0}(y_1))\bigg)^\top\nonumber\\
&\times \bigg(\hat{y}_1-y_1-\sum_{i=1}^p c_i \frac{\partial}{\partial y_{1,i}}\log(\xi_{x_0}(y_1))\bigg) \bigg\}\nonumber\\
=&\mathbb{E}\{(\hat{y}_1-y_1)^\top(\hat{y}_1-y_1)\}\nonumber\\
&-2\sum_{i=1}^p\mathbb{E}\bigg\{(\hat{y}_1-y_1)^\top c_i \frac{\partial}{\partial y_{1,i}}\log(\xi_{x_0}(y_1))\bigg\} \nonumber \\
&+\sum_{i=1}^p\sum_{j=1}^pc_i^\top c_j\mathbb{E}\bigg\{\frac{\partial \log(\xi_{x_0}(y_1)) }{\partial y_{1,i}} \frac{\partial \log(\xi_{x_0}(y_1))}{\partial y_{1,j}}\bigg\}. \label{eqn:proof_lower_bound:1}
\end{align}
Notice that
\begin{align*} 
\mathbb{E}\bigg\{(\hat{y}_1-y_1)^\top c_i&\frac{\partial}{\partial y_{1,i}}\log(\xi_{x_0}(y_1))\bigg\}\\
=&-\int_{y_1} y_1^\top  c_i \frac{\partial}{\partial y_{1,i}}\xi_{x_0}(y_1)\mathrm{d}y_1\\
=&c_{i,i}-\int_{y_1} \nabla_{y_1}.((y_1^\top c_i)(\xi_{x_0}(y_1)e_i))\mathrm{d}y_1,
\end{align*}
where the first and the second equalities, respectively, follow from 
\begin{align*}
\mathbb{E}\bigg\{\hat{y}_1^\top c_i&\frac{\partial}{\partial y_{1,i}}\log(\xi_{x_0}(y_1))\bigg\}\\
&=\int_{y_1}\hat{y}_1^\top c_i\frac{\partial}{\partial y_{1,i}}\log(\xi_{x_0}(y_1))\xi_{x_0}(y_1)\mathrm{d}y_1\\
&=\int_{y_1}\hat{y}_1^\top c_i\frac{\partial \xi_{x_0}(y_1)}{\partial y_{1,i}}\mathrm{d}y_1\\
&=\hat{y}_1^\top c_i\int_{y_1}\bigg[\frac{\partial \xi_{x_0}(y_1)}{\partial y_{1}}^\top e_i\bigg]\mathrm{d}y_1\\
&=0, \hspace{0.8in} \mbox{because of Assumption~\ref{assum:2}~(\textit{ii})}
\end{align*}
and
\begin{align*}
\nabla_{y_1}.((y_1^\top c_i)(\xi_{x_0}(y_1)e_i))=
&(y_1^\top c_i)\nabla_{y_1}.(\xi_{x_0}(y_1)e_i)\\
&+(\xi_{x_0}(y_1)e_i)^\top \nabla_{y_1}(y_1^\top c_i)\\
=&(y_1^\top c_i)\frac{\partial}{\partial y_{1,i}}\xi_{x_0}(y_1)+\xi_{x_0}(y_1)c_{i,i}.
\end{align*}
Here, $e_\ell$ denotes a vector of all zeros except the $\ell$-th entry that is equal to one.
Now, it can be deduced that
$\int_{y_1}\nabla_{y_1}.((y_1^\top c_i)(\xi_{x_0}(y_1)e_i))\mathrm{d}y_1=0,$
because of the Gauss's Theorem (a.k.a., the divergence theorem)~\cite{larson2016calculus} and $\lim_{y_1\rightarrow \infty}y_{i,j}\xi_{x_0}(y_1)=0$ for all~$j$ because $\mathbb{E}\{y_1\}$ is bounded in light of Assumption~\ref{assum:2}~(\textit{iii}). Therefore,
it can be seen that
\begin{align}  \label{eqn:proof_lower_bound:2}
\mathbb{E}\bigg\{&(\hat{y}_1-y_1)^\top c_i\frac{\partial}{\partial y_{1,i}}\log(\xi_{x_0}(y_1))\bigg)\bigg\}=c_{i,i}.
\end{align}
On the other hand, by the definition of the Fisher information matrix in~\eqref{eqn:def:Fisherinformation}, it can be deduced that
\begin{align} \label{eqn:proof_lower_bound:3}
\mathbb{E}\bigg\{\frac{\partial}{\partial y_{1,i}}\log(\xi_{x_0}(y_1)) \frac{\partial}{\partial y_{1,j}}\log(\xi_{x_0}(y_1))\bigg\}=\mathcal{I}_{i,j}.
\end{align}
Substituting~\eqref{eqn:proof_lower_bound:2} and~\eqref{eqn:proof_lower_bound:3} into~\eqref{eqn:proof_lower_bound:1} while setting $c_\ell=(1/\mathcal{I}_{\ell,\ell}) e_\ell$, $\forall\ell$, results in
\begin{align}
\mathbb{E}\bigg\{\bigg\|\hat{y}_1-y_1-&\sum_{i=1}^p c_i \frac{\partial}{\partial y_{1,i}}\log(\xi_{x_0}(y_1))\bigg\|_2^2\bigg\}\nonumber\\
&=\mathbb{E}\{\|\hat{y}_1-y_1\|_2^2 \}-\sum_{i=1}^p 1/\mathcal{I}_{i,i}.
\label{eqn:proof_lower_bound:4}
\end{align}
 Noting that the left-hand-side of~\eqref{eqn:proof_lower_bound:4} is always greater than or equal to zero (because it is the expectation of a non-negative random variable), it can be deduced that
$
\mathbb{E}\{\|\hat{y}_1-y_1\|_2^2 \}
\geq \sum_{i=1}^p 1/\mathcal{I}_{i,i}
\geq p^2/(\sum_{i=1}^p\mathcal{I}_{i,i})\geq 1/(\sum_{i=1}^p\mathcal{I}_{i,i}),
$
where the second inequality follows from the Jensen's inequality~\cite[p.\,25]{rockafellar2015convex}, the facts that the mapping $z\mapsto 1/z$ is convex over $\mathbb{R}_{\geq 0}$, and $\mathcal{I}_{i,i}\geq 0$ for all $i$ (since $\mathcal{I}$ is positive semi-definite). This completes the proof.

\section{Proof of Theorem~\ref{main:results}}
\label{proof:main:results}
First, we prove that $\trace(\mathcal{I})$ is convex in $\xi_{x_0}$, $\forall x_0$. Note that $\trace(\mathcal{I})=\int
\nabla_{y_1} \xi_{x_0}(y_1)^\top \nabla_{y_1} \xi_{x_0}(y_1)/\xi_{x_0}(y_1) \mathrm{d}y_1.$
Define $f:\mathbb{R}_{>0}\times\mathbb{R}^p\rightarrow\mathbb{R}$ as $f(x,y)=y^\top y/x$, which is convex over $\mathbb{R}_{> 0}\times\mathbb{R}^p$ because its Hessian is positive semi-definite over its domain.
Let $\xi_1(y_1)$ and $\xi_2(y_1)$ satisfy Assumption~\ref{assum:2} and belongs to $\aleph_{x_0}$. We may define $\xi_{x_0}(y_1)=\alpha_1\xi_1(y_1)+\alpha_2\xi_2(y_1)$ for any $\alpha_1,\alpha_2\in(0,1)$ such that $\alpha_1+\alpha_2=1$. Clearly $\xi_{x_0}(y_1)$ satisfies Assumption~\ref{assum:2} as well. Further, $\sup(\xi_{x_0})\subseteq\supp(\xi_1)\cup \supp(\xi_2)$ and, as a result, $\xi_{x_0}\in\aleph_{x_0}$. For any $y_1\in\supp(\xi_1)\cap \supp (\xi_2)$, where $\supp(\xi):=\{y_1\,|\,\xi_{x_0}(y_1)>0\}$, it can be shown that
\begin{align}
\frac{1}{\xi_{x_0}(y_1)}\nabla_{y_1} \xi_{x_0}(y_1)^\top \nabla_{y_1} \xi_{x_0}(y_1)\nonumber
 &=
f(\xi_{x_0}(y_1),\nabla_{y_1} \xi_{x_0}(y_1)) \nonumber\\
&\leq 
\alpha_1 f(\xi_1(y_1),\nabla_{y_1} \xi_1(y_1))\nonumber\\
&+\alpha_2 f(\xi_2(y_1),\nabla_{y_1} \xi_2(y_1)),
 \label{eqn:proof:convexity:0}
\end{align}
due to convexity of $f$ over $\mathbb{R}_{>0}\times\mathbb{R}^p$. 
Now, note that, if $y_1\in \supp(\gamma_1(\xi_1))\setminus\supp (\xi_2)$, 
\begin{align*}
f&(\xi_{x_0}(y_1),\nabla_{y_1} \xi_{x_0}(y_1))\nonumber\\
&=
f(\alpha_1\xi_1(y_1),\alpha_1\nabla_{y_1} \xi_1(y_1)+\alpha_2\nabla_{y_1} \xi_2(y_1)) \nonumber\\
&\leq 
\alpha_1 f(\alpha_1\xi_1(y_1),\xi_1(y_1))+\alpha_2 f(\alpha_1\xi_1(y_1),\nabla_{y_1} \xi_2(y_1)) \nonumber\\
&=
f(\alpha_1\xi_1(y_1),\xi_1(y_1))+(\alpha_2/\alpha_1) f(\alpha_1\xi_1(y_1),\nabla_{y_1} \xi_2(y_1))\nonumber\\
&\leq M_1,
\end{align*}
for $M_1>0$. Thus, because $\supp(\xi_1)\setminus \supp (\xi_2)\subseteq \supp (\xi_2)^c$, with $\mathcal{A}^c$ denoting the complement of set $\mathcal{A}$, is a zero-measure set (see Assumption~\ref{assum:2}), $\int_{\supp(\xi_1)\setminus \supp (\xi_2)}
f(\xi(y_1|x_0),\nabla_{y_1} \xi_{x_0}(y_1))\mathrm{d}y_1=0$. Similarly,$
\int_{\supp(\xi_2)\setminus \supp (\xi_1)}
f(\xi(y_1|x_0),\nabla_{y_1} \xi_{x_0}(y_1))\mathrm{d}y_1=0.$ 
In light of these identities, the proof of convexity of $\trace(\mathcal{I})$ follows from integrating both sides of~\eqref{eqn:proof:convexity:0}.

Noting that the cost function and the constraint set are convex, the stationarity condition (that the variational derivative is equal to zero) is sufficient for optimality.  Further, if multiple density functions satisfy the sufficiency conditions, they all exhibit the same cost. 

In the rest of the proof, the stationarity condition is rewritten in a simpler form. 
Following the result of~\cite{jeyakumar1990zero}, the Lagrangian can be constructed as
\begin{align*}
\mathcal{L}
=&\int \bigg(
\frac{1}{\xi_{x_0}(y_1)}\nabla_{y_1} \xi_{x_0}(y_1)^\top \nabla_{y_1} \xi_{x_0}(y_1)\\
&\quad +\varrho U_{x_0}(y_1)\xi_{x_0}(y_1)\bigg)\mathrm{d}y_1+\mu\bigg(\hspace{-.03in}-\hspace{-.03in}\int \xi_{x_0}(y_1)\mathrm{d}y_1\hspace{-.03in}+\hspace{-.03in}1\bigg)\nonumber\\
=&\int \bigg(\frac{1}{\xi_{x_0}(y_1)}\nabla_{y_1} \xi_{x_0}(y_1)^\top \nabla_{y_1} \xi_{x_0}(y_1)-\mu\xi_{x_0}(y_1)\\
&\quad +\varrho U_{x_0}(y_1)\xi_{x_0}(y_1)\bigg)\mathrm{d}y_1+\mu,
\end{align*}
where $\mu\in\mathbb{R}$ is the Lagrange multiplier corresponding to the equality constraint $\int\xi_{x_0}(y_1)\mathrm{d}y_1=1$. Using Theorem~5.3 in~\cite[p.\,440]{edwards1973advanced}, it can be seen that the extrema must satisfy
\begin{align*}
\varrho U_{x_0}(y_1)-\mu-&\frac{1}{\xi_{x_0}(y_1)^2}\nabla_{y_1} \xi_{x_0}(y_1)^\top \nabla_{y_1} \xi_{x_0}(y_1)\\
&-2\sum_{i=1}^p \bigg(-\frac{1}{\xi_{x_0}(y_1)^2} \bigg(\frac{\partial \xi_{x_0}(y_1)}{\partial y_{1,i}}\bigg)^2\\
&+\frac{1}{\xi_{x_0}(y_1)}\frac{\partial^2 \xi_{x_0}(y_1)}{\partial y_{1,i}^2}\bigg)=0.
\end{align*}
Introducing the change of variable $\xi_{x_0}(y_1)=u_{x_0}(y_1)^2$ results in
$
U_{x_0}(y_1)-\mu(x_0)-4\nabla^2_{y_1}u_{x_0}(y_1)/u_{x_0}(y_1)=0,  \forall y_1\in\aleph_{x_0}.
$
Now, it can be deduced that $(1/4)(\varrho U_{x_0}(y_1)-\mu)u_{x_0}(y_1)-\nabla^2_{y_1}u_{x_0}(y_1)=0.$
Note that, if $u_{x_0}(y_1)=0$ for some $y_1$, the equality cannot be satisfied with any $\mu\in\mathbb{R}$. 

\section{Single-Integrator Robots} \label{appendixC}
For single integrator robots, $U_{y_0}(y_1)=(y_1-y_0)^\top(y_1-y_0)/T$. With change of variable $r:=\|y_1-y_0\|_2$, we get $U_{x_0}(y_1)=\tilde{U}(r)= r^2/T$. The polar coordinates, in addition to $r$, require angle $\theta\in(-\pi,\pi]$ such that $y_1-y_0=[r\cos(\theta) \; r\sin(\theta)]^\top$. We get $\aleph_{x_0}:=\{y_1\,|\,|\theta|\leq \theta_{\max}\}$. Changing from Cartesian coordinates to the polar coordinates and substituting $u(r,\theta)=R(r)\cos(\omega\theta)$ with $\omega=\pi/(2\theta_{\max})$ into the partial differential equation in Theorem~\ref{main:results} results in
\begin{align*}
\frac{\mathrm{d}^2}{\mathrm{d}r^2}R(r)+\frac{1}{r}\frac{\mathrm{d}}{\mathrm{d}r}R(r)+\frac{1}{4}
(\mu-\varrho \tilde{U}(r))R(r)-\frac{\omega^2}{r^2}R(r)=0.
\end{align*}
Finally, note that the square root of the following density function solves this  partial differential equation:
\begin{align*}
\xi_{x_0}(y_1)=&\frac{\displaystyle (\varrho/T)^{\left(\pi/(4\theta_{\max})+ 1/2\right)}}{\displaystyle \theta_{\max}\Gamma(\pi/(2\theta_{\max}) + 1)2^{\pi/(2\theta_{\max})}}\\
&\times \exp\left(-\frac{1}{2}\sqrt{\frac{\varrho}{T}}r^2\right)r^{\pi/\theta_{\max}}\cos^2\bigg(\frac{\pi}{2\theta_{\max}}\theta\bigg).
\end{align*}

\end{document}